\documentclass{article}
\usepackage{ijcai16}

\usepackage{times}
\usepackage{epsfig}
\usepackage{graphicx}
\usepackage{caption}
\usepackage{subcaption}
\usepackage{amsmath}
\usepackage{amssymb}
\usepackage{xcolor}
\usepackage{url}

\usepackage{algorithm}
\usepackage[noend]{algpseudocode}

\newcommand{\citenp}[1]{\citeauthor{#1} \shortcite{#1}}
\title{Predicting Personal Traits from Facial Images using Convolutional Neural Networks Augmented with Facial Landmark Information}
\author{Yoad Lewenberg{\Large $^1$} \\yoadlew@cs.huji.ac.il
	\And
	Yoram Bachrach{\Large $^2$} \\yobach@microsoft.com
	\And
	Sukrit Shankar{\Large $^3$} \\ss965@cam.ac.uk
	\And Antonio Criminisi{\Large $^2$} \\ antcrim@microsoft.com
	\AND
	\textnormal{
		{\Large $^1$}The Hebrew University of Jerusalem, Israel} \\
	{\Large $^2$}Microsoft Research, Cambridge, United Kingdom \\
	{\Large $^3$}  Machine Intelligence Lab (MIL), Cambridge University}
\begin{document}

\maketitle

\begin{abstract}
  We consider the task of predicting various traits of a person given an image of their face. We estimate both objective traits, such as gender, ethnicity and hair-color; as well as subjective traits, such as the emotion a person expresses or whether he is humorous or attractive. For sizeable experimentation, we contribute a new Face Attributes Dataset (FAD), having roughly 200,000 attribute labels for the above traits, for over 10,000 facial images. 
  
  Due to the recent surge of research on Deep Convolutional Neural Networks (CNNs),  we begin by using a CNN architecture for estimating facial attributes and show that they indeed provide an impressive baseline performance. To further improve performance, we propose a novel approach that incorporates facial landmark information for input images as an additional channel, helping the CNN learn better attribute-specific features so that the landmarks across various training images hold correspondence.  We empirically analyse the performance of our method, showing consistent improvement over the baseline across traits. 
\end{abstract}

\section{Introduction}

Humans find it very easy to determine various traits  of other people, simply by looking at them. Without almost any conscious effort, a glimpse at another person's face is sufficient for us to ascertain their gender, age or ethnicity. We can easily decide whether they are attractive, look funny or are approachable, or determine the emotion they are displaying (for example, whether they appear sad, happy or surprised). As social creatures, making such inference is clearly important to us. Imparting commensurate capabilities to machines is bound to enable very interesting applications \cite{parikh2012relative}. However,  in contrast to the relative ease with which humans infer such personal traits of an individual from their facial image, training a machine to do the same is a challenging task. 

Deep Convolutional Neural Networks (CNNs)~\cite{lecun1998gradient,behnke2003hierarchical,simard2003best} are prominent statistical learning models, which have recently been shown to be very effective for image classification tasks
~\cite{krizhevsky2012imagenet,bengio2007greedy,deng2009imagenet,szegedy2014going}. 
These networks employ several layers of neuron collections in a feed-forward manner, where the individual neurons are tiled in a way so that they respond to overlapping regions in the visual field. As opposed to hand crafted convolution kernel methods~\cite{maini2009study}, the elements of each convolution kernel in CNNs are trained by backpropagation, applied in conjunction with an optimization technique such as Stochastic Gradient Descent (SGD) \cite{lecun1998gradient}. 

Analyzing facial images has been a key research area in computer vision and artificial intelligence for quite a long time.
Researchers have proposed automated methods for inferring personal traits of individuals from facial images~\cite{lyons1999automatic}, including gender
~\cite{moghaddam2002learning}, 
age
~\cite{horng2001classification} and ethnicity~\cite{lu2004ethnicity,hosoi2004ethnicity}. 

Earlier work has even uncovered methods for predicting more subjective or social traits from facial images~\cite{vinciarelli2009social}, such as the expressed emotion~\cite{padgett1997representing,fasel2003automatic} or attractiveness
\cite{kagian2006humanlike,datta2008algorithmic}.
Very recently, \citenp{MSRGuessMyAge} released mobile and web applications, which were aimed at guessing the age of humans by just looking at their facial images. 

In addition to specific methods for predicting personal attributes, earlier works have also examined reusable building blocks for facial image processing, for tasks such as detecting faces, pose estimation, face segmentation and facial landmark localization~\cite{huang2007labeled,segundo2010automatic,zhu2012face,uvrivcavr2012detector,zhang2014facial}. Such methods provide additional information about the face of a person, improving the accuracy of many facial classification tasks~\cite{lu2005multimodal,uvrivcavr2012detector}. 

\textbf{Drawbacks in prior methods for face analysis:} Despite the success of previously proposed methods in inferring various personal traits from facial images, most solutions are based upon hand-designed features, and typically suffer from one or more of the following problems:
 (a) They are specifically tailored to a single task at hand \cite{su2013ica,tian2003automatic,tjahyadi2007face,hasan2014experiments}; (b) They are not well scalable to real-world variations in data such as multiple view-points~\cite{dhall2011emotion}; (c) They make use of unautomated pre-processing methods such as hand-labeling of key facial regions \cite{attribute_classifiers}. 

\textbf{Major advantages of deep learning:}  Since deep learning based procedures can \textit{automatically} learn a diverse set of low and high-level representations for the input data, they circumvent the need for building hand-crafted features. Also, since deep nets work directly on input images, there is seldom any need to do unautomated or esoteric preprocessing.

\textbf{Applying Deep Learning to a range of facial attributes:} Given the promise of deep learning and the nature of our problem where we aim to predict attributes ranging from objective to   subjective ones, from a diverse set of facial images, CNNs are an excellent fit for our needs. We thus apply CNNs for predicting the facial attributes.  Previous papers on personal attribute prediction with deep nets have either not focused on facial attributes \cite{Shankar_2015_CVPR}, or have only considered a very restricted set of facial attributes such as emotions \cite{lisetti1998facial,liu2014facial}. Also, where researchers have tried to rank facial attributes for better classification \cite{parikh2012relative,shankar2013semantic}, relative attributes and plausibly subjective supervision are required. 

\textbf{Augmenting CNNs with face alignment information:} 
While training, a CNN is inherently expected to learn features in a way which can correctly tell us about the spatial regions in the images most salient for the prediction of a class. For maximum robustness, these spatial regions should be consistent across all the training images of a given class. For instance, the personal traits exhibited in faces generally correspond to specific facial regions or a combination of them - hair color is mostly captured in the hair region of the face; happiness is specific to the region around lips; while old age can be seen as a combination of features around the forehead, under-eyes and cheeks. Thus, for all training images belonging to the class of hair-color, we would like that the CNN learns features that correspond to the hair regions of the image for prediction. If the CNN predicts white / blonde hair-color by considering the white skin color of a person, we would term that as erroneous. While a human can innately and consistently figure out such structural accordances in an image for a given class, the task is rather difficult for CNNs, more so when the classes are attributes (as against the objects).  Noticing that the faces have a well-defined structure (forehead, eyes, noses, mouth, etc) which can be robustly captured using state-of-the-art techniques like \cite{zhang2014facial}, we augment the input data with this structural information to train a CNN. We thus expect it to learn more robust attribute-specific features, thereby ameliorating the prediction accuracy.\footnote{This can also be seen as  a knowledge-transfer approach with deep learning. (Though in a different sense from transfer learning and multi-task learning methods employed with some deep nets \cite{zhou2014hybrid,oquab2014learning,zhanglearning}.)}


\subsection{Our Contribution:}
We contribute a new Face Attributes Dataset (FAD), comprising of roughly 200,000 attribute labels for over 10,000 facial images. Our dataset covers many traits of individuals, and has labels regarding both objective and subjective personal attributes. The dataset has been carefully crowd-sourced from Amazon Mechanical Turk, establishing the veracity of the labels obtained. 

We apply deep learning for predicting a wide range of facial attributes. We corroborate that using a CNN architecture for determining facial attributes provides an impressive baseline performance. To further improve performance, we propose an augmentation approach that incorporates facial landmark information for input images as an additional channel, helping the CNN learn better attribute-specific features so that the landmarks across various training images hold correspondence.  We empirically show consistent improvement with our proposed approach over the aforementioned baseline across traits. 




\section{Face Attributes Dataset (FAD)}
\label{l_sect_data_fad}

Our dataset consists of 10,000 facial images of celebrities (public figures), where each image is tagged with various traits of the individual. The images we used are a subset of the PubFig dataset~\cite{attribute_classifiers}. 

The original PubFig dataset consisted of 60,000 images of celebrities, where each celebrity is covered by multiple images under different poses, at different times, and with a different expression. Due to copyright issues, original images were never provided for the PubFig Dataset, and only the respective internet addresses (URLs) were given. Since the release of PubFig, many of those URLs have become invalid, so we focused on the subset of images of the original data which are still available online. 

The resolution of the 10,000 images downloaded was not constant. Since typically all the input images to a CNN are of the same size, we scaled each image to a fixed resolution of 150 $\times$ 150 pixels. We chose this resolution since most images posted on social media sites do not contain faces bigger than   that (typically people pose with their torsos as well, if not the full body). Our dataset has thus been curated keeping practical applications in mind; so algorithms performing well on our dataset should also perform well on other real-world data. 

\subsection{Ground-Truth Annotations}

As our target variables, we focused on multiple objective and subjective traits; the objective traits include: gender, ethnicity, age, make-up and hair color; the subjective traits include emotional expression, attractiveness, humorousness and chubbiness. The classes considered for each of these traits / attributes are listed in Table~\ref{l_tab_traits} along with their level of skewness. We emphasize that in this paper we consider the prediction of classes for each trait to be a discrete classification problem. (E.g. we only aim to know whether a person's gender is male or female and not the degree to which they appear to be  masculine or feminine.) 

In order to get the images labelled for various traits, we used Amazon's Mechanical Turk (MTurk). This is a crowdsourcing platform, which allows people to post micro-tasks, and lets participants fulfill these tasks for a fee. We sourced a total of 1,500 raters from MTurk. All the participants were sourced from the US and Canada. We let each of the participants examine several images and provide labels for each image for each of the traits listed in Table~\ref{l_tab_traits}. 

\begin{table}
	\centering
	\begin{tabular}{ | c | c | }
		\hline
		Trait & Data distribution \\
		\hline
		\hline
		Gender & Male (50.8\%), Female (49.2\%) \\
		\hline
		Ethnicity & White (79.5\%), Other (20.5\%) \\
		\hline
		Hair Color & Dark (60\%), Bright (40.0\%) \\
		\hline
		Makeup & Wears (39.4\%), Does not wear (60.6\%) \\
		\hline
		Age & Young (67.8\%), Elder (32.2\%) \\
		\hline
		\hline
		Emotions & Joy (64.2\%), Other (35.8\%) \\
		\hline
		Attractive  &  Yes (65.9\%), No (34.1\%) \\
		\hline
		Humorous &  Yes (55.6\%), No (44.4\%) \\
		\hline
		Chubby &   Yes (57.3\%), No (42.7\%) \\
		\hline
	\end{tabular}
	\caption{ \textbf{Attributes / Traits in FAD: }    Personal traits in FAD along with the corresponding classes are listed. For each trait, the distribution of images across the corresponding classes is given. As is evident, some traits have more skewness across their classes as compared to others. For all our experiments, our training and test sets contain a similar distribution. }
	\label{l_tab_traits}
\end{table}

We offered each MTurk participant a payment of \$6 for filling in all the trait labels for 10 of our images. To account for the fact that all participants on MTurk might not exert enough effort (or be satisfactorily sincere) in the annotation task, we made sure we have enough non-redundant labels, by having each image labelled 3 times. 

To further ensure the quality of the labels, we  included some very simple questions designed to identify participants who could be randomly clicking answers or not paying enough attention to the task.\footnote{For example, we asked simple mathematical questions for which every participant is expected to know the answer, such as ``how much is 6+8?''}
We removed the responses of participants who failed to correctly answer these questions. Also, we excluded the responses of participants who disagreed with their peers on over a third of the labels for the objective traits (e.g. participants who did not agree with their peers on the gender or ethnicity labels for a third of their images). 

Our goal is to use the annotations of the images in FAD to train an automated system to infer personal traits from facial images. 
However, some traits are clearly more difficult than others. 
When people find it easy to infer a certain property from an image, we expect a high degree of agreement between the raters. In contrast, when inferring a target variable is difficult, we expect our annotators to often disagree regarding the correct label for an image. 

In cases where a trait exhibits a low degree of inter-rater agreement in the dataset, even an excellent learning method would find it difficult to achieve a high degree of accuracy in the task. Table~\ref{l_tab_interrater_agreement} presents the inter-rater agreement, as measured by Fleiss'  Kappa~\cite{fleiss2013statistical}, for each of the traits,  evaluated on our dataset. 

\begin{table}
	\centering
	\begin{tabular}{ | c | c | }
		\hline
		Trait & Data distribution \\ 
		\hline
		\hline
		Gender & 0.9601  (APA)\\
		\hline
		Ethnicity & 0.913 (APA) \\
		
		\hline 
		Hair Color & 0.719 (SA)\\
		\hline
		Makeup & 0.697 (SA) \\
		\hline
		Age & 0.563  (MA)\\
		\hline
		\hline
		Emotions &0.688 (SA) \\
		\hline
		Attractive  & 0.29 (FA)\\
		\hline
		Humorous & 0.171 (SLA)\\
		\hline
		Chubby & 0.153 (SLA)\\
		\hline
	\end{tabular}
	\caption{\textbf{Inter-rater agreement (Fleiss' Kappa) measured for each of the traits in FAD.}  A value of $1$ indicates perfect agreement, while a value of $0$ indicates no agreement. APA stands for \emph{Almost perfect agreement}; MA for \emph{Moderate agreement}; SA for \emph{Substantial agreement}; FA for \emph{Fair agreement}; and SLA for \emph{Slight agreement}. 
	}
	\label{l_tab_interrater_agreement}
\end{table}

\section{Prediction Algorithm}
\label{l_sec_prediction_algorithm}

\begin{figure}
	\begin{center}
		\includegraphics[width=1.0\columnwidth]{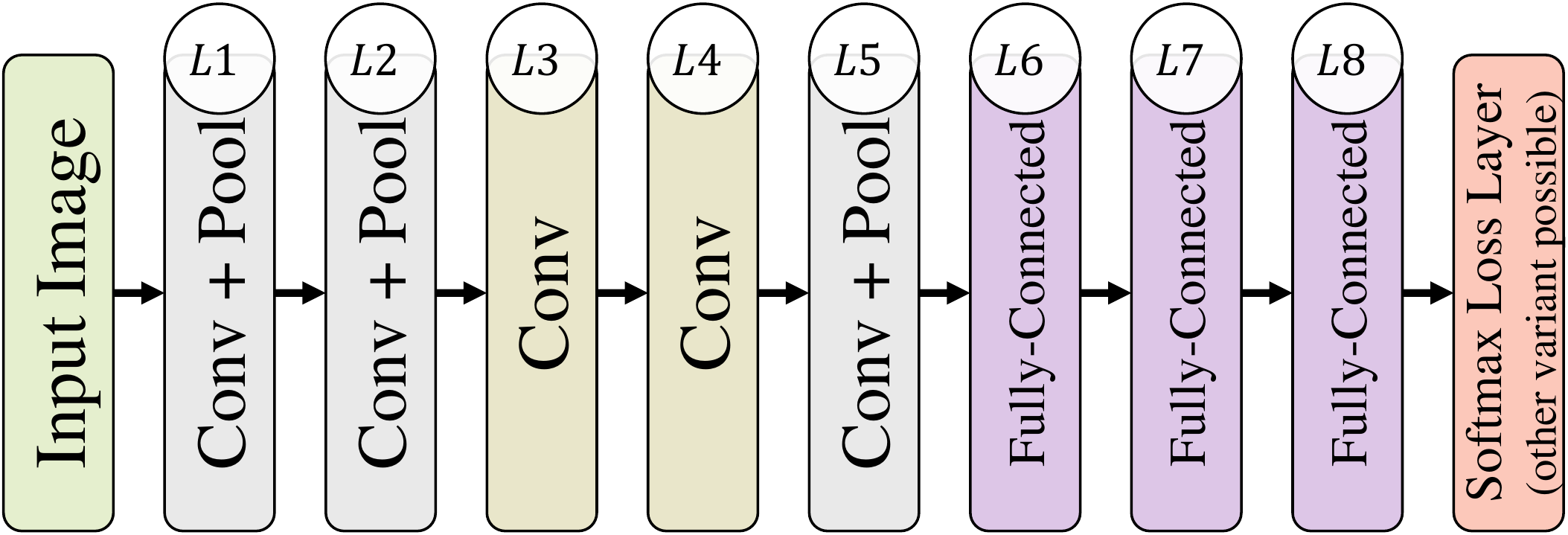}
	\end{center}
	\caption{
		\textbf{Block Illustration of AlexNet \protect \cite{krizhevsky2012imagenet}: }
		a deep CNN architecture. 
		The deep convolutional neural net has eight layers (denoted as $L1,\ldots,L8$) after the input. The last fully connected layer is conventionally followed by a softmax loss layer, but can also be replaced by the likes of Sigmoid Cross Entropy Loss Layer
		\protect \cite{jia2014caffe}. We use this CNN architecture for all our experiments.
	}
	\label{fig_alexnet}
\end{figure}

Our goal is to predict the various traits of a person from their facial images. We consider FAD for all our experiments, and use the traits and the corresponding classes as listed  in Table~\ref{l_tab_traits}.  Due to the various advantages offered by deep nets, we begin by applying one of the most famous CNN architectures \cite{krizhevsky2012imagenet} for our prediction task, widely known as AlexNet. The block-level architecture of AlexNet is shown in Figure~\ref{fig_alexnet}. 

\textbf{Brief Overview of AlexNet:} The fully-connected layers have $4096$ neurons each. Max-pooling is done to facilitate local translation-invariance and dimension reduction. For the fully connected layers, a drop-out \cite{srivastava2014dropout} probability of $0.5$ is used to avoid overfitting. The final fully connected layer  takes the outputs of $L7$ as its input, produces outputs equal to the number of classes through a fully connected architecture, then passes these outputs through a softmax function, and finally applies the negative log likelihood loss. With softmax loss layer, each input image is expected to have only one label. When the softmax loss layer is replaced by a sigmoid cross-entropy loss layer, the outputs of $L8$ are applied to a sigmoid function to produce predicted probabilities, using which a cross-entropy loss is computed. Here each input can have multiple label probabilities. We refer the reader to \cite{krizhevsky2012imagenet} for complete details of AlexNet. 


\textbf{Choice of the Loss Function:} Softmax Loss and Sigmoid Cross-Entropy Loss are the two most widely used loss functions for classification tasks in deep learning. With the softmax loss layer, the training of the AlexNet is  typically accomplished by minimizing the following cost or error function (negative log-likelihood):
\small
\begin{equation}
\mathcal{L}_s = -\dfrac{1}{N} \sum_{r=1}^{N}\, \log (\hat{p}_{r,y_r}) + \mathcal{L}_R 
\end{equation}
\normalsize
where $r$ indexes $N$ training images across all traits ($r \in \{1, \ldots, N\}$), $\mathcal{L}_R  = \lambda || \boldsymbol{W} ||_2$ is the L2 regularization on weights $\boldsymbol{W}$ of the deep net, $\lambda$ is a regularization parameter, and the probability $\hat{p}_{r,y_r}$ is obtained by applying the softmax function to the $M$ outputs of layer $L8$, $M$ being the number of classes we wish to predict labels for. Letting $l_{r,m}$ denote the $m^{th}$ output for $r^{th}$ image, we have
\small
\begin{equation}
\hat{p}_{r,m} = \dfrac{\mathrm{e}^{l_{r,m}}}{\sum_{m'} \mathrm{e}^{l_{r,m'}}}, \qquad ~~m,m' \in \{1, \dots, M\}.
\end{equation}
\normalsize
In case one applies the sigmoid cross entropy loss, each image is expected to be annotated with a vector of ground-truth label probabilities $\boldsymbol{p}_r$, having length $M$, and the network is trained by minimizing the following loss objective:
\small
\begin{equation}
\mathcal{L}_e =  -\dfrac{1}{NM} \sum_{r=1}^{N} \sum_{m=1}^{M} \left[\boldsymbol{p_r} \log (\boldsymbol{\hat{p}_r})  + (1 - \boldsymbol{p_r}) \log (1 -\boldsymbol{\hat{p}_r})\right]   +  \mathcal{L}_R
\end{equation}
\normalsize
\noindent where the probability vector $\boldsymbol{\hat{p}_r}$ is obtained by applying the sigmoid function to each of the $M$ outputs of layer $L8$. 

A natural choice to approach our prediction task is to train a single CNN for all our traits / attributes. Since an image can have multiple traits, the sigmoid cross-entropy loss function is best suited for our scenario. However, we find that for our prediction task where for every given trait, we have mutually exclusive attribute classes, training one net for each given trait provides a higher accuracy. We thus establish our baseline and perform all our experiments with the latter choice. For a greater number of facial traits, one can combine the features from these independently trained CNNs, to train some fully connected layers. 
\begin{figure}[t]
	\centering
	\includegraphics[width=0.95\columnwidth]{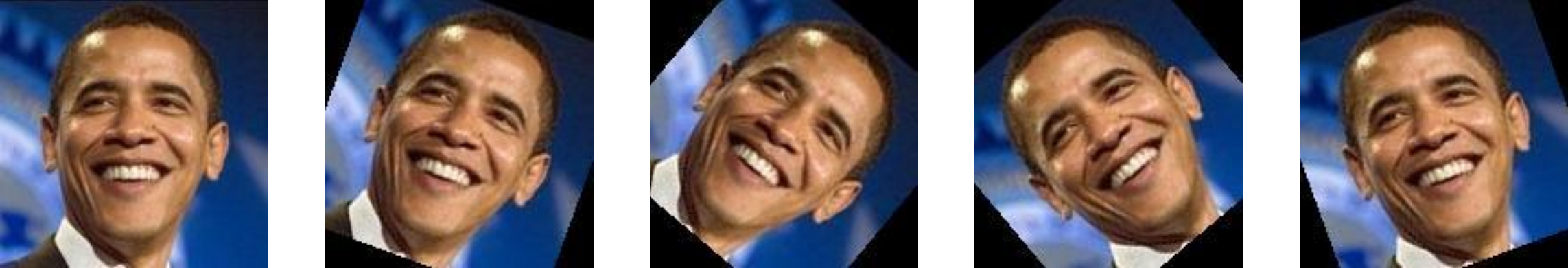}
	\caption{\textbf{Rotation of images in the training set:} In order to make the deep net training more robust to facial pose variation, for each training image (left most), we create 4 new training images as its rotated versions (last four). The original training image is rotated by $\{-40,-20,20,40\}$ degrees.}
	\label{fig:rot}
\end{figure}

\textbf{Rotating images in the training set:} In order to make the deep net training more robust to facial pose variation, for each training image, we create 4 new training images as its rotated versions.  Each training image is rotated by $\{-40,-20,20,40\}$ degrees. This also increases the size of our training set by a factor of 5. Example of the rotated versions of an input image is shown in Figure~\ref{fig:rot}.  

\subsection{Incorporating Facial Landmark Information} 
\label{l_sec_incorporating_facial_landmark}

For each of the traits, we train the network on FAD using labels for the classes of that trait, and evaluate its performance as a baseline. The input to the network is a color image, in a resolution of $D_x \times D_y$ (we used $D_x = D_y = 150$ pixels); each pixel is represented as a three ``channel'' RGB encoding. Thus the input layer has $3$ neurons, each neuron representing a 2-D matrix of size $D_x \times D_y$. 

We propose an improvement to the basic deep convolutional neural network, by
incorporating facial landmark information in the input data. 
Localizing facial landmarks, sometimes referred to as ``face alignment'' is a key step in many facial image analysis approaches. 

Various recognition algorithms, including those dealing with facial figures, require exact positioning of an object into a canonical pose, to allow examining the position of features relative to a fixed coordinate system. Inspired by such methods, we embed landmark information in deep nets for predicting a wide range of facial attributes.

\begin{figure}[t]
	\begin{centering}
		{\includegraphics[width=0.2\columnwidth]{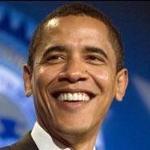}}
		{\includegraphics[width=0.2\columnwidth]{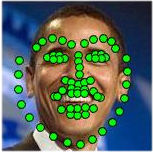}} \\ 
		{\includegraphics[width=0.9\columnwidth]{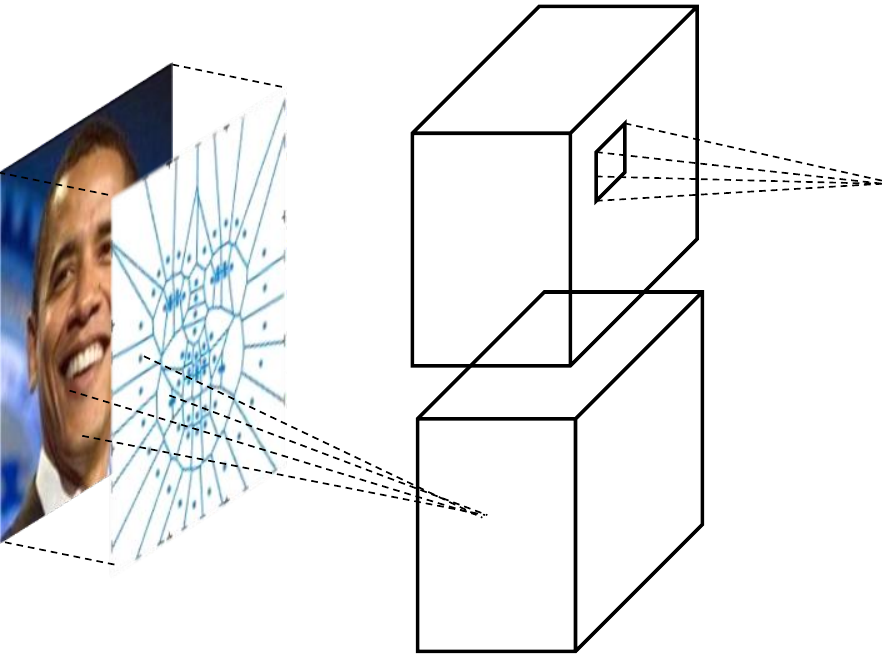}}
		\caption{ 
			\textbf{Using facial landmarks for an input image:} \textbf{(\textit{Top})} Example of an input  image and the corresponding facial landmarks detected using  TCDCN \protect \cite{zhang2014facial}.     \textbf{(\textit{Bottom})}  An illustration of the augmented input channel $A^{(x,y)}$ computed using AUGMENT-FLL. Each region in $A^{(x,y)}$ is coded with a different value as per the index of the associated landmark. The figure shows the additional input channel being fed into the subsequent parts of convolutional neural network, along with the RGB image. \label{fig:flm} 
		}
	\end{centering}	
\end{figure}

Facial landmark localization algorithms are designed to find the location of several key ``landmarks'' in an image, such as the location of the center of the eyes, parts of the nose or the sides of the mouth.  Consider a list $L = (l_1, \ldots, l_k)$ of facial landmarks. Facial landmark localization algorithms receive a facial image $I$ as an input, and output the coordinates in the image for each of the landmarks $C^I = (c^I_1, \ldots, c^I_k)$ where $c^I_j = (x^I_j,y^I_j)$ are the coordinates of landmark $l_j$ in the image $I$. 
An example of an image and the corresponding facial landmarks is given  in Figure~\ref{fig:flm}.

Our improved approach uses a facial landmark localization algorithm as a subroutine, so any such algorithm could be used by our approach. It operates by associating each pixel in the facial image with the {\em closest} facial landmark for that image. We then add this association as an additional channel to each input image. 

We now formally describe our approach. In our baseline approach, the pixel in coordinate $(x,y)$ in the input image $I$ is encoded as three RGB channels $(R^{(x,y)}, G^{(x,y)}, B^{(x,y)})$. We add an additional channel, relating to the closest facial landmark, denoted as $A^{(x,y)}$, thus increasing the number of neurons in the input layer from $3$ to $4$. $A^{(x,y)}$ encodes the identity of the nearest facial landmark to the pixel in coordinates $(x,y)$. 

To compute $A^{(x,y)}$, we call the facial landmark localization algorithm (FLL) as a subroutine, to obtain a list of landmark coordinates, $C^I = (c^I_1, \ldots, c^I_k)$ where $c^I_j = (x^I_j,y^I_j)$, and compute the distance between pixel $(x,y)$ to each of these coordinates, to obtain $d^I_j(x,y) = || (x^I_j,y^I_j), (x,y)  ||_2$. 

We select the index of the facial landmark nearest to the pixel as the value of the pixel in the additional channel. Finally, we train the CNN on the set of augmented images, consisting of the original RGB channels and the new channel encoding the nearest landmark associated with each pixel. We refer to our approach as the \textbf{L}andmark \textbf{A}ugmented \textbf{C}onvolutional \textbf{N}eural \textbf{N}etwork (LACNN) method. 

The algorithm for generating the additional channel is given in Figure~\ref{algo:vor}. For facial landmark detection, we have used the state-of-the-art TCDCN face alignment tool~\cite{zhang2014facial}, which returns the locations of $k=68$ key facial landmarks. In AUGMENT-FLL, TCDCN  is thus used for FLL. We find that TCDCN is fairly robust to facial viewpoint variation. Instead of TCDCN, any other facial landmark detection tool could also be used. An illustration of the augmented input channel $A^{(x,y)}$ is shown in Figure~\ref{fig:flm}.

\begin{figure}

	\begin{algorithmic}
		\Procedure{AUGMENT-FLL\\}{$I = (R^{(x,y)}, G^{(x,y)}, B^{(x,y)})$} 
		\State $(c^I_1, \ldots, c^I_k)$  = $FLL(I)$     \textcolor{gray}{// Get facial landmarks} 
		\For{$x=1$ {\bf to} $D_x$}
		\For{$y=1$ {\bf to} $D_y$}
		\For{$j=1$ {\bf to} $k$}
		\State $d^I_j(x,y) = || (x^I_j,y^I_j), (x,y)  ||_2$
		\State \textcolor{gray}{// pixel-landmark distances} 
		\EndFor
		\State $A^{(x,y)} = \arg \min_{j \in \{1, \ldots, k\}} d^I_j(x,y)$
		\EndFor
		\EndFor
		\State \textbf{return}  $I' = (R^{(x,y)}, G^{(x,y)}, B^{(x,y)}, A^{(x,y)})$ 
		\EndProcedure
	\end{algorithmic}	
	\caption{ \textbf{Creating $A^{(x,y)}$:} Algorithm for encoding the identity of the nearest facial landmark to  every pixel. This algorithm is used to create the additional channel $A^{(x,y)}$ which augments the input images. \label{algo:vor}}
\end{figure}

\section{Results}
\label{l_sect_results}
\begin{table}[t]
	\centering
	\begin{tabular}{ | c | c | c | }
		\hline
		Trait & Baseline & LACNN \\
		\hline
		\hline
		Gender & \bf{98.46\%} & 98.33\%  \\
		\hline
		Ethnicity & 82.7\% & \bf{83.35\%} \\
		\hline		
		Hair Color & 91\% & \bf{91.69\%} \\
		\hline			
		Makeup &  92.5\% & \bf{92.87\%} \\
		\hline
		Age & 88.42\% & \bf{88.83\%} \\			
		\hline
		\hline
		Emotions & \bf{88.93\%} & 88.33\%  \\
		\hline		
		Attractive  & 78.44\% & \bf{78.85\%} \\
		\hline
		Humorous  & 66.8\% & \bf{69.06\%} \\
		\hline
		Chubby & 60.6\% & \bf{61.38\%} \\		
		\hline
	\end{tabular}
	\caption{\textbf{Comparison of prediction accuracy:} The accuracy of the baseline CNN method and LACNN on FAD. For most of the objective and subjective traits, LACNN improves the prediction accuracy. }
	\label{tab:res}
\end{table}

We now discuss the performance of the baseline CNN approach and LACNN. As mentioned before, we use FAD for all our experiments. 

For training and inference with CNNs, we have used the Caffe Library~\cite{jia2014caffe}. For doing inference on a considerable amount of test images, we create a 80/20 train/test split with FAD, maintaining the same data distribution across the training and test sets for all traits as given in Table~\ref{l_tab_traits}. Such a split evaluates our method on roughly 36,000 labels.

Table~\ref{tab:res} shows the accuracy of the baseline CNN method and LACNN on FAD. It is clear that CNN provides an overall impressive baseline performance. 
Even for highly subjective traits, where human raters tend to disagree regarding the correct label of an image (see Table~\ref{l_tab_interrater_agreement}), CNNs give a reasonable performance. This indicates that CNN based approaches are indeed flexible, and can handle many traits without resorting to building ad-hoc systems relying on hand-crafted features.

\begin{figure*}
	\centering		
	\begin{subfigure}[b]{0.1\textwidth}
		\includegraphics[width=\textwidth]{barack_obama_181.jpg}
		\caption{Input}
		\label{fig:graph1}
	\end{subfigure}						
	\begin{subfigure}[b]{0.39\textwidth}
		\includegraphics[width=\textwidth]{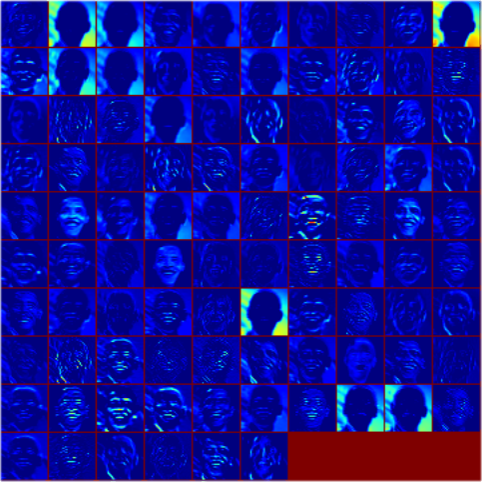}
		\caption{Responses using baseline CNN}
		\label{fig:graph2} 
	\end{subfigure}			
	\begin{subfigure}[b]{0.39\textwidth}
		\includegraphics[width=\textwidth]{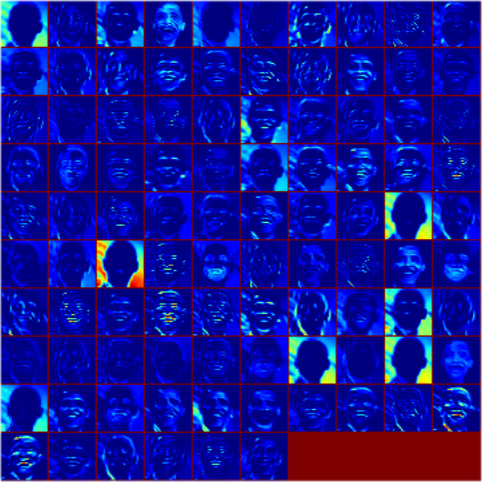}
		\caption{Responses using LACNN }
		\label{fig:graph4}
	\end{subfigure}		
	\caption{\textbf{Visualizing output of the first convolutional layer of AlexNet trained with baseline CNN and LACNN:} For an input image, figure shows the outputs responses of the first convolutional layer of AlexNet trained using the baseline CNN method and our proposed LACNN. Both visualizations have been generated from nets trained on the same trait. The first convolutional layer of AlexNet contains 96 neurons, whose outputs are shown here on a 10 $\times$ 10 grid.  A careful observation reveals that the responses generated using LACNN contain more detailed information as compared to the ones generated with baseline CNN: neurons in LACNN contain more discernible information about key facial parts. Also, more neurons exhibit valuable information in LACNN. 
		\label{fig:lacnn}}

\end{figure*}

The proposed LACNN shows consistent improvement across most of the traits as compared to the CNN baseline. Note that LACNN has the capability to improve performance for both the objective as well as the subjective traits.  This substantiates our intuition that face alignment information can be useful in predicting facial attributes using deep nets. 

To further validate our intuition that facial landmarks should help the CNN learn more robust attribute-specific features in a more consustent manner, we depict the visualizations of the output responses of the first convolutional layer of AlexNet, trained with both baseline CNN and LACNN in Figure~\ref{fig:lacnn}. 

Observing the filter activations of the first convolutional layer shows that the responses generated using LACNN have more detailed information as compared to the ones generated with the baseline of a non-augmented CNN. 

The output responses generated with LACNN have many variations of prominent facial parts including the nose, eyes, hairline, etc. Further, there is a higher number of neurons exhibiting such valuable information in the case of LACNN. 

The outputs with better discernible information can be attributed to the fact that landmark augmentation helps the CNN to learn filters in a way such that similar regions across a range of facial images hold correspondence  to exhibit similar responses. This is clearly important for facial attribute prediction, since a given trait in any face is always associated with the same combination of facial sub-parts.

\section{Conclusions and Future Work}
\label{l_conc}
We have  proposed a method for predicting personal attributes from facial images, based on a CNN architecture augmented with face alignment information. We have empirically evaluated our approach by building a tagged facial images dataset called FAD, showing that improved classification performance can be achieved for a very wide range of traits using our approach. 

Observing the filter activations of the first convolutional layer (Figure~\ref{fig:lacnn})  shows that  the responses generated using LACNN have more detailed, 
which suggests our  technique would be  more robust to image noise.future work could test this hypothesis.

Several questions remain open for further research.  First, could one devise a method using facial landmarks to better detect facial attributes, such that attribute-specific regions are explicitly learned for faces? Could such a method be used for {\em ranking} facial images according to attributes? 

Further, could one detect more subjective attributes such as more detailed emotions or traits such as being in shape (muscle tone) or other health related traits or friendliness? Could such an analysis be based on the information contained in the nets trained for the basic objective attributes? 

Finally, could one exploit graph-structured compositions within the deep nets to better interpret facial traits? More generally, a key disadvantage of CNN based methods is that the learned model is not ``human interpretable'' in the sense that it is difficult to understand which sub-parts of the network drive the prediction. Would it be possible to train multiple nets or a single net for many traits and examine the correlation, so that it would be possible to explain the predictions made by the system in a way understandable by humans?

\small
\bibliographystyle{named}
\bibliography{FaceCNN}

\begin{thebibliography}{}

\bibitem[\protect\citeauthoryear{Behnke}{2003}]{behnke2003hierarchical}
Sven Behnke.
\newblock {\em Hierarchical neural networks for image interpretation}.
\newblock Science \& Business Media, 2003.

\bibitem[\protect\citeauthoryear{Bengio \bgroup \em et al.\egroup
  }{2007}]{bengio2007greedy}
Yoshua Bengio, Pascal Lamblin, Dan Popovici, Hugo Larochelle, et~al.
\newblock Greedy layer-wise training of deep networks.
\newblock {\em NIPS}, 2007.

\bibitem[\protect\citeauthoryear{Datta \bgroup \em et al.\egroup
  }{2008}]{datta2008algorithmic}
Ritendra Datta, Jia Li, and James~Z Wang.
\newblock Algorithmic inferencing of aesthetics and emotion in natural images:
  An exposition.
\newblock In {\em ICIP}, pages 105--108. IEEE, 2008.

\bibitem[\protect\citeauthoryear{Deng \bgroup \em et al.\egroup
  }{2009}]{deng2009imagenet}
Jia Deng, Wei Dong, Richard Socher, Li-Jia Li, Kai Li, and Li~Fei-Fei.
\newblock Imagenet: A large-scale hierarchical image database.
\newblock In {\em CVPR}, pages 248--255. IEEE, 2009.

\bibitem[\protect\citeauthoryear{Dhall \bgroup \em et al.\egroup
  }{2011}]{dhall2011emotion}
Abhinav Dhall, Akshay Asthana, Roland Goecke, and Tom Gedeon.
\newblock Emotion recognition using phog and lpq features.
\newblock In {\em FG}, 2011.

\bibitem[\protect\citeauthoryear{Fasel and Luettin}{2003}]{fasel2003automatic}
Beat Fasel and Juergen Luettin.
\newblock Automatic facial expression analysis: a survey.
\newblock {\em Pattern recognition}, 36(1):259--275, 2003.

\bibitem[\protect\citeauthoryear{Fleiss \bgroup \em et al.\egroup
  }{2013}]{fleiss2013statistical}
Joseph~L Fleiss, Bruce Levin, and Myunghee~Cho Paik.
\newblock {\em Statistical methods for rates and proportions}.
\newblock John Wiley \& Sons, 2013.

\bibitem[\protect\citeauthoryear{Hasan and Pal}{2014}]{hasan2014experiments}
Md~Kamrul Hasan and Christopher Pal.
\newblock Experiments on visual information extraction with the faces of
  wikipedia.
\newblock In {\em AAAI}, 2014.

\bibitem[\protect\citeauthoryear{Horng \bgroup \em et al.\egroup
  }{2001}]{horng2001classification}
Wen-Bing Horng, Cheng-Ping Lee, and Chun-Wen Chen.
\newblock Classification of age groups based on facial features.
\newblock {\em TJSE}, 2001.

\bibitem[\protect\citeauthoryear{Hosoi \bgroup \em et al.\egroup
  }{2004}]{hosoi2004ethnicity}
Satoshi Hosoi, Erina Takikawa, and Masato Kawade.
\newblock Ethnicity estimation with facial images.
\newblock In {\em AFGR}, 2004.

\bibitem[\protect\citeauthoryear{Huang \bgroup \em et al.\egroup
  }{2007}]{huang2007labeled}
Gary~B Huang, Manu Ramesh, Tamara Berg, and Erik Learned-Miller.
\newblock Labeled faces in the wild: A database for studying face recognition
  in unconstrained environments.
\newblock 2007.

\bibitem[\protect\citeauthoryear{Jia \bgroup \em et al.\egroup
  }{2014}]{jia2014caffe}
Yangqing Jia, Evan Shelhamer, Jeff Donahue, Sergey Karayev, Jonathan Long, Ross
  Girshick, Sergio Guadarrama, and Trevor Darrell.
\newblock Caffe: Convolutional architecture for fast feature embedding.
\newblock In {\em ACM Multimedia}, pages 675--678. ACM, 2014.

\bibitem[\protect\citeauthoryear{Kagian \bgroup \em et al.\egroup
  }{2006}]{kagian2006humanlike}
Amit Kagian, Gideon Dror, Tommer Leyvand, Daniel Cohen-Or, and Eytan Ruppin.
\newblock A humanlike predictor of facial attractiveness.
\newblock In {\em NIPS}, pages 649--656, 2006.

\bibitem[\protect\citeauthoryear{Krizhevsky \bgroup \em et al.\egroup
  }{2012}]{krizhevsky2012imagenet}
Alex Krizhevsky, Ilya Sutskever, and Geoffrey~E Hinton.
\newblock Imagenet classification with deep convolutional neural networks.
\newblock In {\em NIPS}, pages 1097--1105, 2012.

\bibitem[\protect\citeauthoryear{Kumar \bgroup \em et al.\egroup
  }{2009}]{attribute_classifiers}
N.~Kumar, A.~C. Berg, P.~N. Belhumeur, and S.~K. Nayar.
\newblock {A}ttribute and {S}imile {C}lassifiers for {F}ace {V}erification.
\newblock In {\em ICCV}. IEEE, Oct 2009.

\bibitem[\protect\citeauthoryear{LeCun \bgroup \em et al.\egroup
  }{1998}]{lecun1998gradient}
Yann LeCun, L{\'e}on Bottou, Yoshua Bengio, and Patrick Haffner.
\newblock Gradient-based learning applied to document recognition.
\newblock {\em Proceedings of the IEEE}, 86(11):2278--2324, 1998.

\bibitem[\protect\citeauthoryear{Lisetti and
  Rumelhart}{1998}]{lisetti1998facial}
Christine~L Lisetti and David~E Rumelhart.
\newblock Facial expression recognition using a neural network.
\newblock In {\em FLAIRS}, 1998.

\bibitem[\protect\citeauthoryear{Liu \bgroup \em et al.\egroup
  }{2014}]{liu2014facial}
Ping Liu, Shizhong Han, Zibo Meng, and Yan Tong.
\newblock Facial expression recognition via a boosted deep belief network.
\newblock In {\em CVPR}, pages 1805--1812. IEEE, 2014.

\bibitem[\protect\citeauthoryear{Lu and Jain}{2004}]{lu2004ethnicity}
Xiaoguang Lu and Anil~K Jain.
\newblock Ethnicity identification from face images.
\newblock In {\em Defense and Security}, 2004.

\bibitem[\protect\citeauthoryear{Lu \bgroup \em et al.\egroup
  }{2005}]{lu2005multimodal}
Xiaoguang Lu, Hong Chen, and Anil~K Jain.
\newblock Multimodal facial gender and ethnicity identification.
\newblock In {\em Adv. Biometrics}. 2005.

\bibitem[\protect\citeauthoryear{Lyons \bgroup \em et al.\egroup
  }{1999}]{lyons1999automatic}
Michael~J Lyons, Julien Budynek, and Shigeru Akamatsu.
\newblock Automatic classification of single facial images.
\newblock {\em TPAMI}, 1999.

\bibitem[\protect\citeauthoryear{Maini and Aggarwal}{2009}]{maini2009study}
Raman Maini and Himanshu Aggarwal.
\newblock Study and comparison of various image edge detection techniques.
\newblock {\em IJIP}, 2009.

\bibitem[\protect\citeauthoryear{Microsoft}{2015}]{MSRGuessMyAge}
Microsoft.
\newblock {Guess my Age}.
\newblock
  \url{https://www.microsoft.com/en-us/store/apps/guess-my-age/9nblggh3t1ld},
  2015.

\bibitem[\protect\citeauthoryear{Moghaddam and
  Yang}{2002}]{moghaddam2002learning}
Baback Moghaddam and Ming-Husan Yang.
\newblock Learning gender with support faces.
\newblock {\em TPAMI}, 24(5):707--711, 2002.

\bibitem[\protect\citeauthoryear{Oquab \bgroup \em et al.\egroup
  }{2014}]{oquab2014learning}
Maxime Oquab, Leon Bottou, Ivan Laptev, and Josef Sivic.
\newblock Learning and transferring mid-level image representations using
  convolutional neural networks.
\newblock In {\em CVPR}, pages 1717--1724. IEEE, 2014.

\bibitem[\protect\citeauthoryear{Padgett and
  Cottrell}{1997}]{padgett1997representing}
Curtis Padgett and Garrison~W Cottrell.
\newblock Representing face images for emotion classification.
\newblock {\em NIPS}, pages 894--900, 1997.

\bibitem[\protect\citeauthoryear{Parikh \bgroup \em et al.\egroup
  }{2012}]{parikh2012relative}
Devi Parikh, Adriana Kovashka, Amar Parkash, and Kristen Grauman.
\newblock Relative attributes for enhanced human-machine communication.
\newblock In {\em AAAI}, 2012.

\bibitem[\protect\citeauthoryear{Segundo \bgroup \em et al.\egroup
  }{2010}]{segundo2010automatic}
Maur{\'\i}cio~Pamplona Segundo, Luciano Silva, Olga Regina~Pereira Bellon, and
  Chau{\~a}C Queirolo.
\newblock Automatic face segmentation and facial landmark detection in range
  images.
\newblock {\em Systems, Man, and Cybernetics}, 2010.

\bibitem[\protect\citeauthoryear{Shankar \bgroup \em et al.\egroup
  }{2013}]{shankar2013semantic}
Sukrit Shankar, Joan Lasenby, and Roberto Cipolla.
\newblock Semantic transform: Weakly supervised semantic inference for relating
  visual attributes.
\newblock In {\em ICCV}, 2013.

\bibitem[\protect\citeauthoryear{Shankar \bgroup \em et al.\egroup
  }{2015}]{Shankar_2015_CVPR}
Sukrit Shankar, Vikas~K. Garg, and Roberto Cipolla.
\newblock Deep-carving: Discovering visual attributes by carving deep neural
  nets.
\newblock In {\em CVPR}, June 2015.

\bibitem[\protect\citeauthoryear{Simard \bgroup \em et al.\egroup
  }{2003}]{simard2003best}
Patrice~Y Simard, Dave Steinkraus, and John~C Platt.
\newblock Best practices for convolutional neural networks applied to visual
  document analysis.
\newblock In {\em ICDAR}, 2003.

\bibitem[\protect\citeauthoryear{Srivastava \bgroup \em et al.\egroup
  }{2014}]{srivastava2014dropout}
Nitish Srivastava, Geoffrey Hinton, Alex Krizhevsky, Ilya Sutskever, and Ruslan
  Salakhutdinov.
\newblock Dropout: A simple way to prevent neural networks from overfitting.
\newblock {\em JMLR}, 15(1):1929--1958, 2014.

\bibitem[\protect\citeauthoryear{Su \bgroup \em et al.\egroup
  }{2013}]{su2013ica}
Roger Su, Timothy~Michael Dockins, and Manfred Huber.
\newblock Ica analysis of face color for health applications.
\newblock In {\em FLAIRS}, 2013.

\bibitem[\protect\citeauthoryear{Szegedy \bgroup \em et al.\egroup
  }{2014}]{szegedy2014going}
Christian Szegedy, Wei Liu, Yangqing Jia, Pierre Sermanet, Scott Reed, Dragomir
  Anguelov, Dumitru Erhan, Vincent Vanhoucke, and Andrew Rabinovich.
\newblock Going deeper with convolutions.
\newblock 2014.

\bibitem[\protect\citeauthoryear{Tian and Bolle}{2003}]{tian2003automatic}
Ying-li Tian and Ruud~M Bolle.
\newblock Automatic detecting neutral face for face authentication and facial
  expression analysis.
\newblock In {\em Intelligent Multimedia Knowledge Management}, 2003.

\bibitem[\protect\citeauthoryear{Tjahyadi \bgroup \em et al.\egroup
  }{2007}]{tjahyadi2007face}
Ronny Tjahyadi, Wanquan Liu, Senjian An, and Svetha Venkatesh.
\newblock Face recognition via the overlapping energy histogram.
\newblock In {\em IJCAI}, 2007.

\bibitem[\protect\citeauthoryear{U{\v{r}}i{\v{c}}{\'a}{\v{r}} \bgroup \em et
  al.\egroup }{2012}]{uvrivcavr2012detector}
Michal U{\v{r}}i{\v{c}}{\'a}{\v{r}}, Vojt{\v{e}}ch Franc, and V{\'a}clav
  Hlav{\'a}{\v{c}}.
\newblock Detector of facial landmarks learned by the structured output svm.
\newblock {\em VISAPP}, 2012.

\bibitem[\protect\citeauthoryear{Vinciarelli \bgroup \em et al.\egroup
  }{2009}]{vinciarelli2009social}
Alessandro Vinciarelli, Maja Pantic, and Herv{\'e} Bourlard.
\newblock Social signal processing: Survey of an emerging domain.
\newblock {\em IVC}, 2009.

\bibitem[\protect\citeauthoryear{Zhang \bgroup \em et al.\egroup
  }{2014}]{zhang2014facial}
Zhanpeng Zhang, Ping Luo, Chen~Change Loy, and Xiaoou Tang.
\newblock Facial landmark detection by deep multi-task learning.
\newblock In {\em ECCV}. 2014.

\bibitem[\protect\citeauthoryear{Zhang \bgroup \em et al.\egroup
  }{2015}]{zhanglearning}
Zhanpeng Zhang, Ping Luo, Chen~Change Loy, and Xiaoou Tang.
\newblock Learning deep representation for face alignment with auxiliary
  attributes.
\newblock {\em TPAMI}, 2015.

\bibitem[\protect\citeauthoryear{Zhou \bgroup \em et al.\egroup
  }{2014}]{zhou2014hybrid}
Joey~Tianyi Zhou, Sinno~Jialin Pan, Ivor~W Tsang, and Yan Yan.
\newblock Hybrid heterogeneous transfer learning through deep learning.
\newblock In {\em AAAI}, 2014.

\bibitem[\protect\citeauthoryear{Zhu and Ramanan}{2012}]{zhu2012face}
Xiangxin Zhu and Deva Ramanan.
\newblock Face detection, pose estimation, and landmark localization in the
  wild.
\newblock In {\em CVPR}, pages 2879--2886. IEEE, 2012.

\end{thebibliography}
\pagebreak

\end{document}